# Data Mining: A Prediction for Performance Improvement of Engineering Students using Classification

Surjeet Kumar Yadav
Research Scholar,
Shri Venkateshwara University
J.P. Nagar, India

Saurabh Pal
Head, Dept. Of MCA,
VBS Purvanchal University
Jaunpur, India

*Abstract*—Now-a-days the amount of data stored in educational database increasing rapidly. These databases contain hidden information for improvement of students' performance. Educational data mining is used to study the data available in the educational field and bring out the hidden knowledge from it. Classification methods like decision trees, Bayesian network etc can be applied on the educational data for predicting the student's performance in examination. This prediction will help to identify the weak students and help them to score better marks. The C4.5, ID3 and CART decision tree algorithms are applied on engineering student's data to predict their performance in the final exam. The outcome of the decision tree predicted the number of students who are likely to pass, fail or promoted to next year. The results provide steps to improve the performance of the students who were predicted to fail or promoted. After the declaration of the results in the final examination the marks obtained by the students are fed into the system and the results were analyzed for the next session. The comparative analysis of the results states that the prediction has helped the weaker students to improve and brought out betterment in the result.

## I. INTRODUCTION

Data mining concepts and methods can be applied in various fields like marketing, medicine, real estate, customer relationship management, engineering, web mining etc. Educational data mining is a new emerging technique of data mining that can be applied on the data related to the field of education. There are increasing research interests in using data mining in education. This new emerging field, called Educational Data Mining, concerns with developing methods that discover knowledge from data originating from educational environments. Educational Data Mining uses many techniques such as Decision Trees, Neural Networks, Naïve Bayes, K- Nearest neighbor, and many others.

Using these techniques many kinds of knowledge can be discovered such as association rules, classifications and clustering. The discovered knowledge can be used for prediction regarding enrolment of students in a particular course, alienation of traditional classroom teaching model, detection of unfair means used in online examination, detection of abnormal values in the result sheets of the students, prediction about students" performance and so on.

Examination plays a vital role in any student's life. The marks obtained by the student in the examination decide his future. Therefore it becomes essential to predict whether the student will pass or fail in the examination. If the prediction says that a student tends to fail in the examination prior to the examination then extra efforts can be taken to improve his studies and help him to pass the examination.

In this connection, the objectives of the present investigation were framed so as to assist the low academic achievers in engineering and they are:

a) Generation of a data source of predictive variables,

b) Identification of different factors, which affects a student's learning behavior and performance during academic career,

c) Construction of a prediction model using classification data mining techniques on the basis of identified predictive variables and

d) Validation of the developed model for engineering students studying in Indian Universities or Institutions.

## II. DECISION TREE

A decision tree is a flow-chart-like tree structure, where each internal node is denoted by rectangles, and leaf nodes are denoted by ovals. All internal nodes have two or more child nodes. All internal nodes contain splits, which test the value of an expression of the attributes. Arcs from an internal node to its children are labeled with distinct outcomes of the test. Each leaf node has a class label associated with it.

Decision tree are commonly used for gaining information for the purpose of decision -making. Decision tree starts with a root node on which it is for users to take actions. From this node, users split each node recursively according to decision tree learning algorithm. The final result is a decision tree in which each branch represents a possible scenario of decision and its outcome.

The three widely used decision tree learning algorithms are: ID3, C4.5 and CART.

### A. ID3 (Iterative Dichotomiser 3)

This is a decision tree algorithm introduced in 1986 by Quinlan Ross [1]. It is based on Hunts algorithm. The tree is constructed in two phases. The two phases are tree building and pruning.

ID3 uses information gain measure to choose the splitting attribute. It only accepts categorical attributes in building a tree model. It does not give accurate result when there is noise. To remove the noise pre-processing technique has to be used.

To build decision tree, information gain is calculated for each and every attribute and select the attribute with the highest information gain to designate as a root node. Label the attribute as a root node and the possible values of the attribute are represented as arcs. Then all possible outcome instances are tested to check whether they are falling under the same class or not. If all the instances are falling under the same class, the node is represented with single class name, otherwise choose the splitting attribute to classify the instances.

Continuous attributes can be handled using the ID3 algorithm by discretizing or directly, by considering the values to find the best split point by taking a threshold on the attribute values. ID3 does not support pruning.

### B. C4.5

This algorithm is a successor to ID3 developed by Quinlan Ross [2]. It is also based on Hunt's algorithm. C4.5 handles both categorical and continuous attributes to build a decision tree. In order to handle continuous attributes, C4.5 splits the attribute values into two partitions based on the selected threshold such that all the values above the threshold as one child and the remaining as another child. It also handles missing attribute values. C4.5 uses Gain Ratio as an attribute selection measure to build a decision tree. It removes the biasness of information gain when there are many outcome values of an attribute.

At first, calculate the gain ratio of each attribute. The root node will be the attribute whose gain ratio is maximum. C4.5 uses pessimistic pruning to remove unnecessary branches in the decision tree to improve the accuracy of classification.

### C. CART

CART [1] stands for Classification And Regression Trees introduced by Breiman. It is also based on Hunt's algorithm. CART handles both categorical and continuous attributes to build a decision tree. It handles missing values.

CART uses Gini Index as an attribute selection measure to build a decision tree .Unlike ID3 and C4.5 algorithms, CART produces binary splits. Hence, it produces binary trees. Gini Index measure does not use probabilistic assumptions like ID3, C4.5. CART uses cost complexity pruning to remove the unreliable branches from the decision tree to improve the accuracy.

## III. BACKGROUND AND RELATED WORK

Data mining techniques can be used in educational field to enhance our understanding of learning process to focus on identifying, extracting and evaluating variables related to the learning process of students as described by Alaa el-Halees [3]. Mining in educational environment is called Educational Data Mining.

Han and Kamber [4] describes data mining software that allow the users to analyze data from different dimensions, categorize it and summarize the relationships which are identified during the mining process.

Bharadwaj and Pal [5] conducted study on the student performance based by selecting 300 students from 5 different degree college conducting BCA (Bachelor of Computer Application) course of Dr. R. M. L. Awadh University, Faizabad, India. By means of Bayesian classification method on 17 attributes, it was found that the factors like students' grade in senior secondary exam, living location, medium of teaching, mother's qualification, students other habit, family annual income and student's family status were highly correlated with the student academic performance.

Pandey and Pal [6] conducted study on the student performance based by selecting 600 students from different colleges of Dr. R. M. L. Awadh University, Faizabad, India. By means of Bayes Classification on category, language and background qualification, it was found that whether new comer students will performer or not.

Hijazi and Naqvi [7] conducted as study on the student performance by selecting a sample of 300 students (225 males, 75 females) from a group of colleges affiliated to Punjab university of Pakistan. The hypothesis that was stated as "Student's attitude towards attendance in class, hours spent in study on daily basis after college, students' family income, students' mother's age and mother's education are significantly related with student performance" was framed. By means of simple linear regression analysis, it was found that the factors like mother's education and student's family income were highly correlated with the student academic performance.

Khan [8] conducted a performance study on 400 students comprising 200 boys and 200 girls selected from the senior secondary school of Aligarh Muslim University, Aligarh, India with a main objective to establish the prognostic value of different measures of cognition, personality and demographic variables for success at higher secondary level in science stream. The selection was based on cluster sampling technique in which the entire population of interest was divided into groups, or clusters, and a random sample of these clusters was selected for further analyses. It was found that girls with high socio-economic status had relatively higher academic achievement in science stream and boys with low socioeconomic status had relatively higher academic achievement in general.

Z. J. Kovacic [9] presented a case study on educational data mining to identify up to what extent the enrolment data can be used to predict student's success. The algorithms CHAID and CART were applied on student enrolment data of information system students of open polytechnic of New Zealand to get two decision trees classifying successful and unsuccessful students. The accuracy obtained with CHAID and CART was 59.4 and 60.5 respectively.

Galit [10] gave a case study that use students data to analyze their learning behavior to predict the results and to warn students at risk before their final exams.

Al-Radaideh, et al [11] applied a decision tree model to predict the final grade of students who studied the C++ course in Yarmouk University, Jordan in the year 2005. Three different classification methods namely ID3, C4.5, and the NaïveBayes were used. The outcome of their results indicated that Decision Tree model had better prediction than other models.

Bharadwaj and Pal [12] obtained the university students data like attendance, class test, seminar and assignment marks from the students' previous database, to predict the performance at the end of the semester.

Ayesha, Mustafa, Sattar and Khan [13] describe the use of k-means clustering algorithm to predict student's learning activities. The information generated after the implementation of data mining technique may be helpful for instructor as well as for students.

Pandey and Pal [14] conducted study on the student performance based by selecting 60 students from a degree college of Dr. R. M. L. Awadh University, Faizabad, India. By means of association rule they find the interestingness of student in opting class teaching language.

Bray [15], in his study on private tutoring and its implications, observed that the percentage of students receiving private tutoring in India was relatively higher than in Malaysia, Singapore, Japan, China and Sri Lanka. It was also observed that there was an enhancement of academic performance with the intensity of private tutoring and this variation of intensity of private tutoring depends on the collective factor namely socioeconomic conditions.

Yadav, Bharadwaj and Pal [16] obtained the university students data like attendance, class test, seminar and assignment marks from the students' previous database, to predict the performance at the end of the semester with the help of three decision trees. It was observed that C4.5 is the best algorithm.

## IV. DATA MINING PROCESS

Predicting the academic outcome of a student needs lots of parameters to be considered. Prediction models that include all personal, social, psychological and other environmental variables are necessitated for the effective prediction of the performance of the students. Data pertaining to student's background knowledge about the subject, the proficiency in attending a question, the ability to complete the examination in time etc will also play a role in predicting his performance.

### A. Data Preparations

The data set used in this study was obtained from VBS Purvanchal University, Jaunpur (Uttar Pradesh) on the sampling method for Institute of Engineering and Technology for session 2010. Initially size of the data is 90.

### B. Data selection and transformation

In this step only those fields were selected which were required for data mining. The data was collected through the enrolment form filled by the student at the time of admission. The student enter their demographic data (category, gender etc), past performance data (SSC or 10th marks, HSC or 10 + 2 exam marks etc.), address and contact number. Most of the attributes reveal the past performance of the students. A few derived variables were selected. While some of the information for the variables was extracted from the database. All the predictor and response variables which were derived from the database are given in Table 1 for reference.

TABLE I
STUDENT RELATED VARIABLES

| Variables | Description | Possible Values |
|---|---|---|
| Branch | Students Branch | {CS, IT, ME} |
| Sex | Students Sex | {Male, Female} |
| Cat | Students category | {Unreserved, OBC, SC, ST} |
| HSG | Students grade in High School | {O – 90% -100%, A – 80% - 89%, B – 70% - 79%, C – 60% - 69%, D – 50% - 59%, E – 40% - 49%, F - < 40%} |
| SSG | Students grade in Senior Secondary | {O – 90% -100%, A – 80% - 89%, B – 70% - 79%, C – 60% - 69%, D – 50% - 59%, E – 40% - 49%, F - < 40% } |
| Atype | Admission Type | {UPSEE, Direct} |
| Med | Medium of Teaching | {Hindi, English} |
| LLoc | Living Location of Student | {Village, Town, Tahseel, District} |
| Hos | Student live in hostel or not | {Yes, No} |
| FSize | student's family size | {1, 2, 3, >3} |
| FStat | Students family status | {Joint, Individual} |
| FAIn | Family annual income | {BPL, poor, medium, high} |
| FQual | Fathers qualification | {no-education, elementary, secondary, UG, PG, Ph.D. NA} |
| MQual | Mother's Qualification | {no-education, elementary, secondary, UG, PG, Ph.D. NA} |
| FOcc | Father's Occupation | {Service, Business, Agriculture, Retired, NA} |
| MOcc | Mother's Occupation | {House-wife (HW), Service, Retired, NA} |
| Result | Result in B. Tech Ist Year | {Pass, Promoted, Fail} |

The domain values for some of the variables were defined for the present investigation as follows:

- Branch – The courses offered by VBS Purvanchal University, Jaunpur are Computer Science and Engineering (CSE), Information Technology (IT) and Mechanical Engineering (ME).

- Cat – From ancient time Indians are divided in many categories. These factors play a direct and indirect role in the daily lives including the education of young people. Admission process in India also includes different percentage of seats reserved for different categories. In terms of social status, the Indian population is grouped into four categories: General, Other Backward Class (OBC), Scheduled Castes (SC) and Scheduled Tribes (ST). Possible values are Unreserved, OBC, SC and ST.

- HSG - Students grade in High School education. Students who are in state board appear for six subjects each carry 100 marks. Grade are assigned to all students using following mapping O – 90% to 100%, A – 80% - 89%, B – 70% - 79%, C – 60% - 69%, D – 50% - 59%, E – 40% - 49%, and F - < 40%.

- SSG - Students grade in Senior Secondary education. Students who are in state board appear for five subjects each carry 100 marks. Grade are assigned to all students using following mapping O – 90% to 100%, A – 80% - 89%, B – 70% - 79%, C – 60% - 69%, D – 50% - 59%, E – 40% - 49%, and F - < 40%.

- Atype - The admission type which may be through Uttar Pradesh State Entrance Examination (UPSEE) or direct admission through University procedure.

- Med – This paper study covers only the colleges of Uttar Pradesh state of India. Here, medium of instructions are Hindi or English.

- FSize-. According to population statistics of India, the average number of children in a family is 3.1. Therefore, the possible range of values is from one to greater than three.

- Result – Students result in Ist year of Engineering. Result is declared as response variable. It is split into three classes Pass, Fail or Promoted. If a student passes all the paper is awarded pass class. If students fail in up to three theory and two practical subjects of an academic year or vice versa, he/she is promoted to next class, otherwise he/she is fail.

C. *Implementation of Mining Model*

Weka is open source software that implements a large collection of machine leaning algorithms and is widely used in data mining applications [11]. From the above data, engg.arff file was created. This file was loaded into WEKA explorer. The classify panel enables the user to apply classification and regression algorithms to the resulting dataset, to estimate the accuracy of the resulting predictive model, and to visualize erroneous predictions, or the model itself. There are 16 decision tree algorithms like ID3, J48, ADT etc. implemented in WEKA. The algorithm used for classification is ID3, C4.5 and CART. Under the "Test options", the 10-fold cross-validation is selected as our evaluation approach. Since there is no separate evaluation data set, this is necessary to get a reasonable idea of accuracy of the generated model. The model is generated in the form of decision tree. These predictive models provide ways to predict whether a new student will perform or not.

D. *Results and Discussion*

The three decision trees as examples of predictive models obtained from the student data set by three machine learning algorithms: the ID3 decision tree algorithm, the C4.5 decision tree algorithm and the CART algorithm.

Figure 1, 2 and 3 shows the rules generated by ID3, C4.5 and CART respectively.

```
SSG = A
| Focc = Service: Pass
| Focc = Business
| | Sex = M: Promoted
| | Sex = F: Pass
| Focc = NA: null
| Focc = Retired: null
| Focc = Agri: Pass
SSG = B
| HSG = A
| | Med  = English: Pass
| | Med  = Hindi: Promoted
| HSG = O: Pass
| HSG = B
| | FAIn = Medium
| | | Atype = UPSEE: Pass
| | | Atype = Direct: Promoted
| | FAIn = High: Fail
| | FAIn = Poor: Fail
| | FAIn = BPL: null
| HSG = C: Promoted
| HSG = E: null
| HSG = D: Promoted
| HSG = F: null
SSG = C
| Fqual = PG
| | Branch = CSE: Pass
| | Branch = IT: Promoted
| | Branch = ME: null
| Fqual = UG
| | HSG = A
| | | Branch = CSE: Promoted
| | | Branch = IT: null
| | | Branch = ME: Pass
| | HSG = O: null
| | HSG = B: Pass
| | HSG = C
| | | Sex = M: Pass
| | | Sex = F: Promoted
| | HSG = E: Promoted
| | HSG = D: Promoted
| | HSG = F: null
| Fqual = Ph.D.: Promoted
| Fqual = Secondary: Fail
| Fqual = Elementary
| | Branch = CSE: null
| | Branch = IT: Pass
| | Branch = ME: Promoted
```

```
       SSG = D
       |  Fqual = PG: null
       |  Fqual = UG
       |  |  Branch = CSE: Fail
       |  |  Branch = IT: Promoted
       |  |  Branch = ME: Promoted
       |  Fqual = Ph.D.: null
       |  Fqual = Secondry: Promoted
       |  Fqual = Elementry: Pass
       SSG = O: Pass
       SSG = E
       |  Cat = Unreserved: null
       |  Cat = OBC: Fail
       |  Cat = SC: Fail
       |  Cat = ST: Promoted
       SSG = F: Promoted
```

Figure 1: ID3 Rules

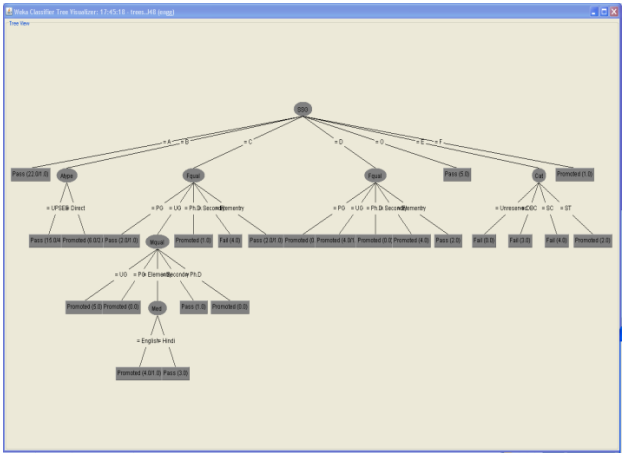

Figure 2: C4.5 Rules

```
SSG=(O)|(A): Pass(26.0/1.0)
SSG!=(O)|(A)
|  Focc=(Service)|(Business)
|  |  FAIn=(Poor)|(High)
|  |  |  Lloc=(Village)|(Town)|(Tahseel):
Promoted(11.0/0.0)
|  |  |  Lloc!=(Village)|(Town)|(Tahseel):
Fail(4.0/1.0)
|  |  FAIn!=(Poor)|(High)
|  |  |  Mocc=(Service)
|  |  |  |  Branch=(ME): Pass(2.0/0.0)
|  |  |  |  Branch!=(ME): Promoted(7.0/0.0)
|  |  |  Mocc!=(Service)
|  |  |  |  HSG=(O)|(B)|(A)|(E)|(F):
Pass(15.0/1.0)
|  |  |  |  HSG!=(O)|(B)|(A)|(E)|(F)
|  |  |  |  |  Sex=(F): Promoted(2.0/0.0)
|  |  |  |  |  Sex!=(F): Pass(2.0/1.0)
|  Focc!=(Service)|(Business)
|  |  HSG=(A): Promoted(3.0/0.0)
|  |  HSG!=(A)
|  |  |  Sex=(F): Promoted(2.0/1.0)
|  |  |  Sex!=(F): Fail(9.0/2.0)
```

Figure 3: CART Decision Tree

The Table II shows the accuracy of ID3, C4.5 and CART algorithms for classification applied on the above data sets using 10-fold cross validation is observed as follows:

TABLE II: CLASSIFIERS ACCURACY

| Algorithm | Correctly Classified Instances | Incorrectly Classified Instances |
|---|---|---|
| ID3 | 62.2222% | 26.6667% |
| C4.5 | 67.7778% | 32.2222 % |
| CART | 62.2222% | 37.7778% |

Table II shows that a C4.5 technique has highest accuracy of 67.7778% compared to other methods. ID3 and CART algorithms also showed an acceptable level of accuracy.

The Table III shows the time complexity in seconds of various classifiers to build the model for training data.

TABLE III: EXECUTION TIME TO BUILD THE MODEL

| Algorithm | Execution Time (Sec) |
|---|---|
| ID3 | 0.00 |
| C4.5 | 0.03 |
| CART | 0.09 |

Table IV below shows the three machine learning algorithms that produce predictive models with the best class wise accuracy.

TABLE IV: CLASSIFIERS ACCURACY

| Algorithm | Class | TP Rate | FP Rate |
|---|---|---|---|
| ID3 | Pass | 0.714 | 0.184 |
| | Promoted | 0.625 | 0.232 |
| | Fail | 0.786 | 0.061 |
| C4.5 | Pass | 0.745 | 0.209 |
| | Promoted | 0.517 | 0.213 |
| | Fail | 0.786 | 0.092 |
| CART | Pass | 0.809 | 0.349 |
| | Promoted | 0.31 | 0.18 |
| | Fail | 0.643 | 0.105 |

## V.  CONCLUSIONS

One of the data mining techniques i.e., classification is an interesting topic to the researchers as it is accurately and efficiently classifies the data for knowledge discovery. Decision trees are so popular because they produce classification rules that are easy to interpret than other classification methods. Frequently used decision tree classifiers are studied and the experiments are conducted to find the best classifier for prediction of student's performance in First Year of engineering exam. From the classifiers accuracy it is clear that the true positive rate of the model for the FAIL class is 0.786 for ID3 and C4.5 decision trees that means model is successfully identifying the students who are likely to fail. These students can be considered for proper counseling so as to improve their result.

Machine learning algorithms such as the C4.5 decision tree algorithm can learn effective predictive models from the student data accumulated from the previous years. The empirical results show that we can produce short but accurate prediction list for the student by applying the predictive models to the records of incoming new students. This study will also work to identify those students which needed special

AUTHORS PROFILE

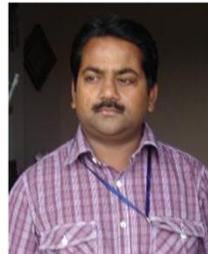

Surjeet Kumar Yadav received his M.Sc. (Computer Science) from Dr. Baba Sahed Marathwada University, Aurangabad, Maharastra, India (1998). At present, he is working as Sr. Lecturer at Department of Computer Applications, VBS Purvanchal Uniersity, Jaunpur. He is an active member of CSI and National Science Congress. He is currently doing research in Data Mining and Knowledge Discovery. He has publisher three research papers.

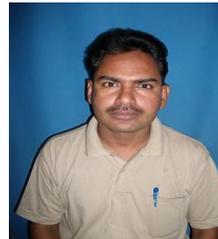

Saurabh Pal received his M.Sc. (Computer Science) from Allahabad University, UP, India (1996) and obtained his Ph.D. degree from the Dr. R. M. L. Awadh University, Faizabad (2002). He then joined the Dept. of Computer Applications, VBS Purvanchal University, Jaunpur as Lecturer. At present, he is working as Head and Sr. Lecturer at Department of Computer Applications.

Saurabh Pal has authored more than 25 research papers in international/national Conference/journals and also guides research scholars in Computer Science/Applications. He is an active member of CSI, Society of Statistics and Computer Applications and working as reviewer for more than 15 international journals. His research interests include Image Processing, Data Mining, Grid Computing and Artificial Intelligence.